\pdfoutput=1

\documentclass[11pt]{article}

\usepackage{amsmath}
\usepackage{booktabs}
\usepackage{multirow}
\usepackage{graphicx}

\usepackage{acl}
\usepackage{times}
\usepackage{latexsym}

\usepackage[T1]{fontenc}

\usepackage[utf8]{inputenc}

\usepackage{microtype}

\title{\textbf{SPELL}: \textbf{S}emantic \textbf{P}rompt \textbf{E}volution based on a \textbf{LL}M\\\textcolor{cyan}{ \normalsize{Discussion and comparison with contemporaneous works in Appendix \ref{appendix-cont}}}}

\author{Yujian Betterest Li$^{\dagger\diamondsuit}$\thanks{\quad Correspondence: \url{bebetterest@outlook.com}} \quad  Kai Wu$^{\dagger}$\\
  $^\dagger$School of Artificial Intelligence, Xidian University \\
  $^\diamondsuit$Institute of Freedom and Happiness, Dream Place\\}
\begin{document}
\maketitle
\begin{abstract}

Prompt engineering is a new paradigm for enhancing the performance of trained neural network models. For optimizing text-style prompts, existing methods usually individually operate small portions of a text step by step, which either breaks the fluency or could not globally adjust a prompt. Since large language models (LLMs) have powerful ability of generating coherent texts token by token, can we utilize LLMs for improving prompts? Based on this motivation, in this paper, considering a trained LLM as a text generator, we attempt to design a black-box evolution algorithm for automatically optimizing texts, namely \textbf{SPELL} (\textbf{S}emantic \textbf{P}rompt \textbf{E}volution based on a \textbf{LL}M). The proposed method is evaluated with different LLMs and evolution parameters in different text tasks. Experimental results show that \textbf{SPELL} could rapidly improve the prompts indeed. We further explore the evolution process and discuss on the limitations, potential possibilities and future work.
\end{abstract}

\begin{figure}[ht]
\small
\centering
\includegraphics[width=1\columnwidth]{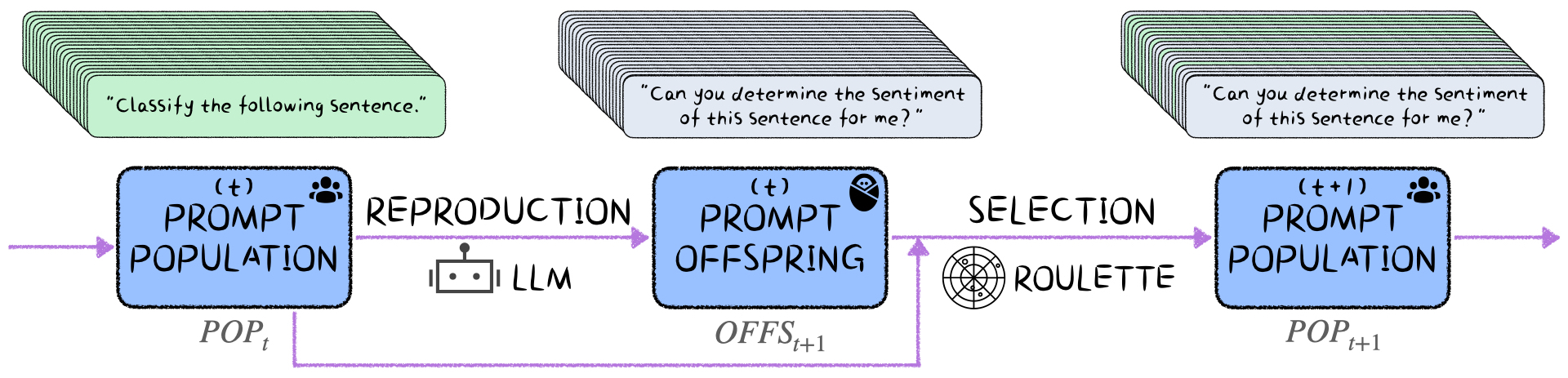}
\caption{Overview of \textbf{SPELL} (\textbf{S}emantic \textbf{P}rompt \textbf{E}volution based on a \textbf{LL}M).}
\label{figure-spell}
\vspace{-0.4cm}
\end{figure}

\section{Introduction}

Prompts (\citealp{liu2023pre}) are additional contents which would be integrated with the original input data to enhance the performance of a trained target model. For example, in a text classification task, a prompt, "[input text]\textbackslash nDivide the above text into 2 classes: positive and negative.\textbackslash nClass:[MASK]", may strengthen the performance of a language model because the prompt reduces the probabilities of many unrelated tokens and hints what the model should focus on. Good prompt engineering methods are able to enhance the ability of a target model with limited and efficient optimization. Besides, prompts are flexible. They could be continuous (\citealp{li-liang-2021-prefix}; \citealp{qin-eisner-2021-learning}), possibly work in hidden layers (\citealp{sun-etal-2022-bbtv2}), and have already been applied to many modalities, such as text (\citealp{lester-etal-2021-power}; \citealp{liu-etal-2022-p}), vision (\citealp{jia2022visual}; \citealp{Sohn_2023_CVPR}), and graph (\citealp{10.1145/3534678.3539249}; \citealp{yi2023contrastive}). In this paper, we only focus on discrete text-style prompts applied at the input of a language model for text tasks.

\begin{figure}[t]
\centering
\includegraphics[width=1\columnwidth]{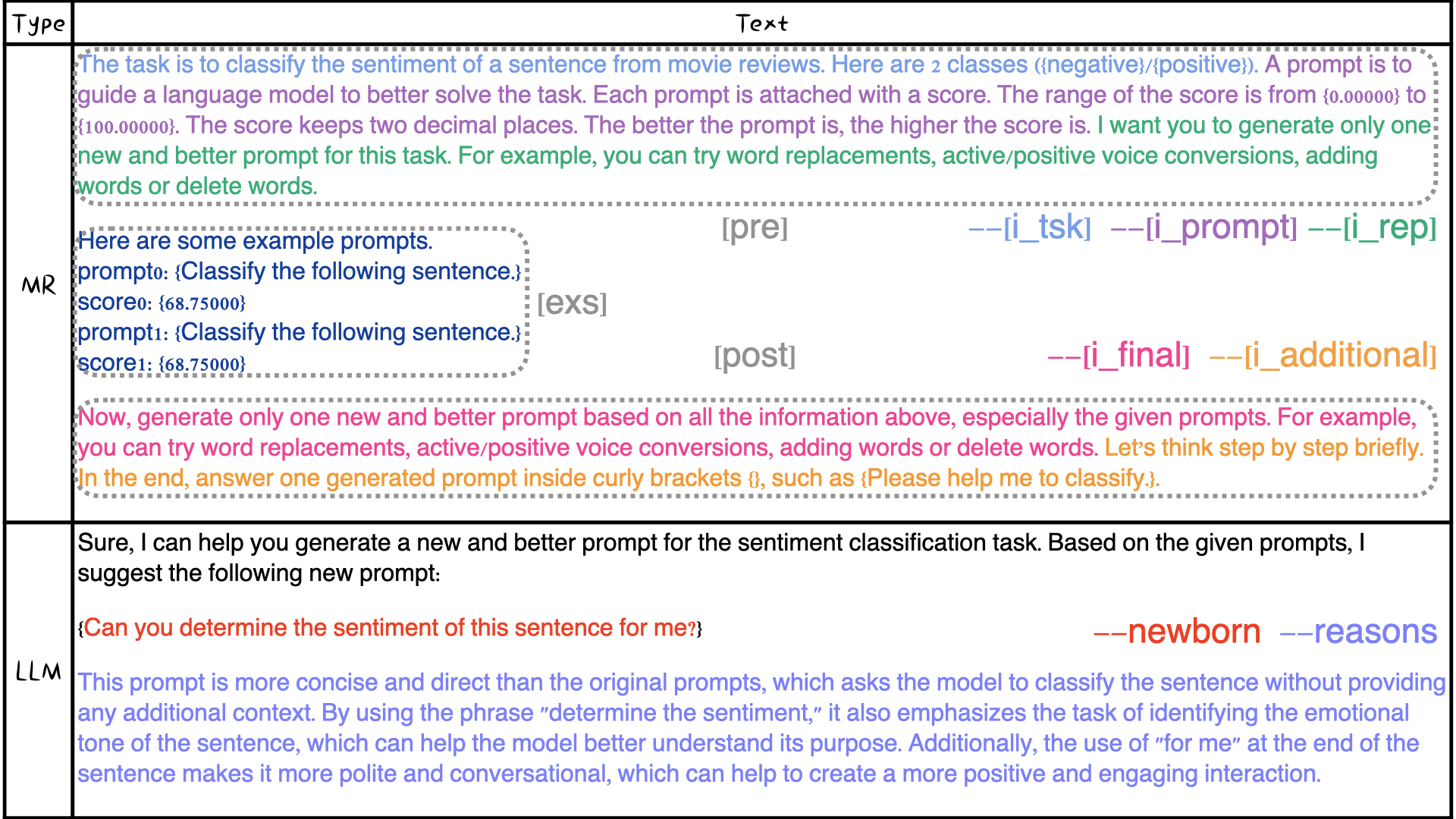}
\caption{A reproduction example including $MP_r$ and the response from the LLM.}
\label{figure-example}
\vspace{-0.4cm}
\end{figure}

There are many methods to optimize text-style prompts. Gradient-based methods (\citealp{wallace-etal-2019-universal}; \citealp{shin-etal-2020-autoprompt}) are the most intuitive one. They usually relax the discrete embeddings for optimization and need large computation because of the process of backward through the trained model. Nowadays, many target models are considered as black boxes since only inference APIs are provided for the users. Hence, many black-box methods, based on evolution (\citealp{pmlr-v162-sun22e}; \citealp{sun-etal-2022-bbtv2}), reinforcement learning (\citealp{deng-etal-2022-rlprompt}; \citealp{zhang2023tempera}; \citealp{diao2023blackbox}) or search strategies (\citealp{prasad-etal-2023-grips}), improve the prompts without the information of gradients and parameters inside the target model. In particular, evolution-based methods maintain a population consists of several prompts as individuals and optimize the population by biological mechanisms such as reproduction and selection. However, these methods individually operate small portions of text step by step, leading to lack of global adjustment and serious damage on the fluency. The prompts optimized by these methods usually consist of various \emph{strange} characters, such as "an-lione\_-lo-:)eak".

In recent years, large language models (LLMs) (\citealp{ijcai2021p0612}; \citealp{zhao2023survey}) are the most popular star among various new AI applications. Optimized on language modeling, they could get universal generalization ability. With appropriate instructions, they could acquire amazing performance on language understanding, reasoning and text generation. Especially, the generated texts are coherent enough.

Therefore, we attempt to propose \textbf{SPELL} (\textbf{S}emantic \textbf{P}rompt \textbf{E}volution based on a \textbf{LL}M), an evolution algorithm to optimize prompts. Since the proposed method considers a LLM as a generator, \textbf{SPELL} would utilize the global feature of a prompt and the improved prompts are coherent. Through experiments, we evaluate that \textbf{SPELL} could rapidly improve performance keeping prompts fluent. Besides, we discover the evolution process and the impacts of different LLMs or evolution strategies. The limitations, challenges and possible directions are discussed on in the end.

\section{Methodology}
As shown in Figure \ref{figure-spell}, \textbf{SPELL} maintains a population $POP_t$ of $N_{POP}$ prompt individuals initialized as $I_{ini}$. By two biological mechanisms, reproduction and selection, the population evolves as
\begin{gather*}
OFFS_t=reproduction(POP_t)\\
POP_{t+1}=selection(OFFS_t \cup POP_t)
\end{gather*}
where $reproduction$ generates the offspring $OFFS_t$ from $POP_t$ and the next population $POP_{t+1}$ consists of individuals selected from the union of $OFFS_t$ and $POP_t$. As evolving for $T$ times, the population is gradually optimized by generating better individuals and selecting the potential ones. In the evolution process, how to inherit outstanding feature and make effective explorations are the keys. In the following subsections, we introduce different components of \textbf{SPELL} and explain how they work in detail.

\subsection{Problem Formulation}
In this paper, we focus on the text tasks translated to classification. For example, sentiment analysis is to classify the emotion contained in a given text. The text $X$ for a target language model $M$ is constructed as 
$$
X=concat([input],\backslash n,[prompt],\backslash n,[head])
$$
where $concat$ stands for concatenate and $[input]$ for the input example. $\backslash n$ is considered as an interval. $[prompt]$ is the prompt to enhance $M$. $[head]$ is set as "class:" to guide $M$ to make the prediction. Next, by language modeling, $M$ makes the prediction as
$$
y=\mathop{\arg\min}\limits_{l_i \in L} P_M(l_i|x_1,x_2...x_{N_X})
$$
where L is the label space. To simplify the prediction process, we choose one word standing for each label, such as "positive" and "negative" in sentiment analysis. It is worth mentioning that we pay attention to optimizing $[prompt]$ keeping $M$ fixed.

\subsection{Reproduction}
\label{section-reproduction}
Reproduction is the process of generating newborns. To keep useful feature from the previous generation, we select $N_P$ individuals $I_{P,t}$ from $POP_t$ as parents for each child, which is introduced in Section \ref{section-selection}. Besides, the newborn individuals generated by a LLM $G$ naturally include random variations as explorations. We construct the core meta prompt for reproduction $MP_{r}$ as
$$
MP_{r}=concat([pre],\backslash n,[exs],\backslash n,[post])
$$
where
\begin{gather*}
\begin{cases}
[pre]=concat([i_{tsk}],[i_{prompt}],[i_{rep}]) \\
[exs]=concat([pro_1],\backslash n,[pro_2]...[pro_{N_P}]) \\
[post]=concat([i_{final}],[i_{additional}]) \\
[pro_i]=concat([I_{P,t,i}],[fit_i]), i \in [1,N_P]
\end{cases}
\end{gather*}
$[pre]$ describes the information on the task ($[i_{tsk}]$), defines what a prompt is for ($[i_{prompt}]$) and leads to reproduction ($[i_{rep}]$). $[exs]$ shows selected parents for reference ($[pro]$). $[post]$ emphasizes generation requests ($[i_{final}]$) and points out other additional instructions ($[i_{additional}]$). Besides, $[fit_i]$ is the fitness/score for the corresponding individual $[I_{P,t,i}]$. By default, $[i_{prompt}]$ is set as "A prompt is to guide a language model to better solve the task. Each prompt is attached with a score. The better the prompt is, the higher the score is."; $[i_{rep}]$ is "I want you to generate only one new and better prompt for this task. For example, you can try word replacements, active/positive voice conversions, adding words or delete words."; $[i_{final}]$ stands for "Now, generate only one new and better prompt based on all the information above, especially the given prompts. For example, you can try word replacements, active/positive voice conversions, adding words or delete words." for detailed generation requests; $[i_{additional}]$ is introduced later and $[i_{task}]$ is listed in Table \ref{table-addition}. Given a meta prompt $MP_{r}$, an individual $I_{o,t,i}$ of the offspring $OFFS_t$ is obtained by 
$$
I_{N,t,i}=EXT(G(MP_{r}))
$$
where $EXT$ extracts the new prompt by the rule included in $[i_{additional}]$. Specifically, in this paper, $[i_{additional}]$ is set as "Let’s think step by step briefly. In the end, output one generated prompt inside curly brackets \{\}, such as \{Please help me to classify.\}.", which means $EXT$ would take out the content inside a pair of curly brackets. In particular, the optimization is prompt-wise global and semantic because $G$ generates texts by language modeling. After that, we obtain the next population $POP_{t+1}$ by selecting individuals from $POP_{t}$ and $OFFS_t$. 

\subsection{Selection}
\label{section-selection}
Selection is to choose some individuals being parents or structuring the new population. Selection strategies usually pay attention to balance survival of the fittest and the luckiest. In \textbf{SPELL}, the selection is based on roulette. First, we calculate the selection probability $P_{S,i}$ for an individual $I_i$ among a candidate set $CAN$ as 
$$
P_{S,i}=\frac{e^{fitness(I_i)}}{\sum_{j \in CAN}(e^{fitness(I_j)})}
$$
where $fitness$ is the metric function which describes how a individual performs in the target task. Better the individual is, higher the corresponding selection probability is, which is helpful for convergence. Then, $N_S$ selected individuals $I_S$ are sampled from $CAN$ as
$$
I_S=sample(CAN,N_S,P_S)
$$
In this way, poor individual would have a chance to be selected, which enriches the diversity.

\section{Experiments}

\paragraph{Setups} We experiment on three text tasks from GLUE (\citealp{wang2018glue}): SST-2 for sentiment prediction, RTE for textual entailment recognition, and AG’s News for news classification. Following \citealp{sun2022black}, our experiments are in the few-shot setting. Only $k$ samples for each class are selected as $k$-shot training data $D_{train}$. The original development data is the test data $D_{test}$, and for datasets without development data, the original test data is used. All datasets and the trained target model are implemented by huggingface\footnote{\url{https://huggingface.co}} and all LLM APIs are available on qianfan platform\footnote{\url{https://cloud.baidu.com/product/wenxinworkshop}}. In addition, by default, the trained target model is RoBERTa-large\footnote{\url{https://huggingface.co/roberta-large}} (\citealp{liu2019roberta}) while the LLM for reproduction is Llama-2-Chat-7b (\citealp{touvron2023llama}); for each round, 5 new individuals are generated based on 1 parent and and another 5 newborns based on 2 parents; elite preservation is implemented for keeping the best individual; $N_{POP}$=$N_s$=20, $T$=500, $k$=16, the seed is set to 42, $fitness$ is the classification accuracy and the maximum $fitness$ of individuals in a population is considered as the performance of the population.

\begin{table}[ht]
\centering
\scriptsize
\begin{tabular}{cccc}
\toprule
Method         & SST-2(\%) & RTE(\%)  & AG’s News(\%)\\ 
\midrule
Zero-shot      & 70.6  & 45.5 & 54.3\\
\midrule
Fine-tuning    & 81.4  & 54.4 & 86.2\\
LM-BFF         & 92.3  & 66.6 & 87.1\\
DART           & 93.5  & 59.0 & 86.8\\ 
\midrule
BBT            & 88.2  & 49.1 & 81.2\\ 
RLPrompt       & 90.5  & 52.2 & 76.2\\
PROMPTBOOSTING & 87.6  & 60.0 & 85.2\\
TEMPERA        & 91.9  & 60.3 & 85.5\\
\midrule
SPELL          & 89.2  & 50.9 & 70.0\\ 
\bottomrule    
\end{tabular}
\caption{Performance of different methods in $D_{test}$. Zero-shot utilizes the initial prompt for prediction. Fine-tuning, LM-BFF (\citealp{gao-etal-2021-making}) and DART (\citealp{zhang2022differentiable}) are white-box methods while BBT (\citealp{pmlr-v162-sun22e}), RLPrompt (\citealp{deng-etal-2022-rlprompt}), PROMPTBOOSTING (\citealp{pmlr-v202-hou23b}), TEMPERA (\citealp{zhang2023tempera}), and SPELL (ours) are black-box methods.}
\label{table-methods}
\vspace{-0.4cm}
\end{table}

\paragraph{Overall View}

We compare \textbf{SPELL} with state-of-the-art methods and feasible black-box works. As shown in Table \ref{table-methods}, as a black-box method, \textbf{SPELL} could improve the accuracy and shows the potential of prompt optimization problems.

\begin{figure}[ht]
\small
\begin{minipage}[t]{0.5\textwidth}
    \centering
    \includegraphics[width=0.8\columnwidth]{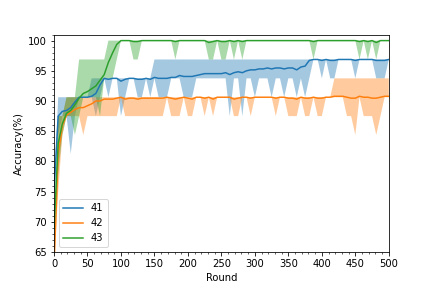}
\end{minipage} 
\\
\begin{minipage}[t]{0.5\textwidth}
    \centering
    \includegraphics[width=0.8\columnwidth]{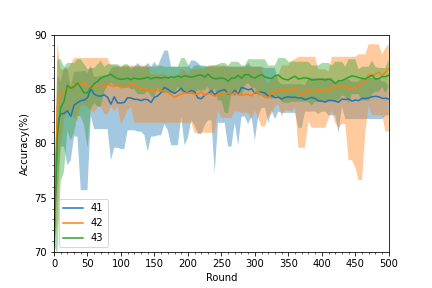}
\end{minipage}
\caption{Accuracy vs. round with different seeds. Three seeds (41, 42 and 43) are considered. The top one is evaluated in $D_{train}$ while the bottom one in $D_{test}$.}
\label{figure-process}
\vspace{-0.4cm}
\end{figure}

\paragraph{Optimization Process} The optimization process is described in Figure \ref{figure-process}. Due to the few-shot setting, there exists a performance gap between $D_{train}$ and $D_{test}$. \textbf{SPELL} optimizes with fewer rounds than many black-box methods, such as BBT and RLPROMPT (more than 3k rounds for prompts of only 5 tokens). In addition, it is worth noting that, due to the randomness of evolution process and text generation, \textbf{SPELL} is \textcolor{red}{\textbf{unstable}} with obvious fluctuations. In addition, the individual generation process is \textcolor{red}{\textbf{semantically causal}} and newborn prompts are \textcolor{red}{\textbf{coherent}} since it is spelled by the LLM token by token according to language modeling. For example, in SST-2, given the initial prompt "Classify the following sentence.", LLM would globally improve it to a new fluent one "Can you determine the sentiment of this sentence for me?" with clear reasons "...more concise and direct...emphasizes the task...more polite and conversational". A reproduction example is shown in Figure \ref{figure-example}.

\begin{table}[ht]
\centering
\scriptsize
\begin{tabular}{cccc}
\toprule
LLM                          & Size(B) & Train(\%) & Test(\%) \\
\midrule
Llama2-Chat                  & 7       & 100.0     & 89.2     \\
Llama2-Chat                  & 13      & 96.9      & 90.0     \\
BLOOMZ                    & 7       & n/a         & n/a        \\
ERNIE-Bot                    & n/a       & 100.0     & 90.7     \\
\bottomrule  
\end{tabular}
\caption{Performance on various LLMs in SST-2. Two Llama2 (\citealp{touvron2023llama}) models with different size are evaluated. BLOOMZ (\citealp{muennighoff2022crosslingual}) fails to generate individuals following instructions. The size of ERNIE-Bot (\citealp{baidu_ernie}) is not available. }
\label{table_llms}
\vspace{-0.4cm}
\end{table}

\paragraph{Influence of Different LLMs}As shown in Table \ref{table_llms}, we evaluate \textbf{SPELL} with different LLMs. Without any especial adaptation, not every LLM could be utilized in \textbf{SPELL}. For example, BLOOMZ is not available because it can not reply following the hinted format. In addition, ERBIE-Bot performs the best while the test accuracy of Llama2-Chat-13B is higher than that of Llama2-Chat-7B. It seems that larger and better LLMs would have better performance. Thus, the LLM for generation is essential for \textbf{SPELL}.

\begin{table}[h]
\centering
\scriptsize
\begin{tabular}{ccc}
\toprule
Configuration & Train(\%) & Test(\%) \\
\midrule
\emph{Base}          & 100.0     & 89.2     \\
\midrule
$fitness$=\emph{loss}  & 71.9      & 69.4     \\
\midrule
$N_{POP}$=2           & 100.0     & 88.9     \\
$N_{POP}$=50          & 100.0     & 87.0     \\
\midrule
$k$=32          & 95.3      & 86.2     \\
$k$=64          & 92.2      & 90.7     \\
\bottomrule  
\end{tabular}
\caption{Results of ablation studies in SST-2. \emph{Base} means the default settings ($fitness$=accuracy; $N_{POP}$=20; $k$=16). \emph{loss} refers to the cross-entropy loss.}
\label{table_ablation}
\vspace{-0.4cm}
\end{table}

\paragraph{Ablation Studies} As shown in Table \ref{table_ablation}, we attempt to find out the impact of some meaningful factors. Even though previous methods usually utilize $loss$ for optimization, we find that \textbf{SPELL} could not work well when $fitness$ is set as $loss$. It may demonstrate that a pure LLM could not effectively exploit the information contained in $loss$. Instead, our method could improve the population simply based on the accuracy. In addition, for the balance of exploration and greediness, the size of the population $N_{POP}$ should be neither too small or large. Besides, larger $D_{train}$ trends to obtain higher accuracy for better estimation.


\section{Conclusion and Discussion}

In this paper, we propose \textbf{SPELL}, a semantic prompt evolution method considering a LLM as a prompt generator. We demonstrate that the proposed method could globally optimize prompts quickly, the reproduction process is reasonable and new spelled prompts are coherent. Besides, we do some ablation studies for further investigation. Even if our method seems unstable, it shows huge latent capacity. Future work may focus on adjusting a LLM to be more adaptive for semantic prompt optimization, such as meta prompt optimization and parameter effective parameter-efficient alignment, stabilizing the optimization, and expanding semantic evolution beyond text optimization.

\bibliography{custom}

\appendix
\section{Other Settings}
\label{appendix-table}

\begin{table}[h]
\small
\begin{tabular}{|c| p{5.8cm}|}
\hline
Task  & \multicolumn{1}{c|}{$i_{task}$}                                                                                               \\ \hline
SST-2 & "The task is to classify the sentiment of a sentence from movie reviews. Here are 2 classes (\{negative\}/\{positive\})." \\ \hline
RTE &
  "The task is to classify the relationship between sentence1 and sentence2. Here are 2 classes (\{yes\}/\{no\}). \{yes\} means the relationship is entailment, the meaning of one sentence is entailed (can be inferred) from the other sentence, otherwise \{no\}." \\ \hline
Task  & \multicolumn{1}{c|}{$I_{ini}$}                                                                                           \\ \hline
SST-2 & "Classify the following sentence."                                                                                        \\ \hline
RTE   & "Classify whether the relationship between the following sentence1 and sentence2 is entailment."                          \\ \hline
\end{tabular}
\caption{A list of $i_{task}$ and $I_{ini}$ in the experiments.}
\label{table-addition}
\vspace{-0.4cm}
\end{table}

\section{On the Contemporaneous Works}
\label{appendix-cont}
We discuss on two recent and similar contemporaneous works which utilize a LLM as a prompt generator/optimizer as well. \textcolor{cyan}{\textbf{It is worth noticing that OPRO and EVOPROMPT perform rather well while SPELL (ours) does not. Moreover, there's overlap between the datasets on which EVOPROMPT is evaluated and those of ours. Considering the difference of the target model, we think the reason may be that the accuracy is originally higer when the target model is a LLM, and that the improved prompts generated by a LLM could not be well "understood" by a relatively small language model.}} In addition, the comparisons of OPRO and EVOPROMPT with SPELL (ours) are illustrated in detail as follows:

\paragraph{OPRO (\citealp{yang2023large})} The target model of OPRO is a LLM while that of SPELL is a much smaller language model. Besides, we think the key difference is that OPRO constructs a prompt-score trajectory of prompts to update the prompts while SPELL (ours) introduces selection and reproduction with randomness from evolution algorithm for prompt improvement.

\paragraph{EVOPROMPT (\citealp{guo2023connecting})} EVOPROMPT and SPELL are really similar. EVOPROMPT includes evolution algorithm as well and further apply differential evolution with a LLM. EVOPROMPT separately designs the crossover and mutation operators while SPELL (ours) consists of one general operator for reproduction. Besides, the target model of EVOPROMPT is a LLM.

\end{document}